%% file: main.tex
\documentclass{article}

\usepackage[preprint,nonatbib]{neurips_2019}

\usepackage[preprint,nonatbib]{}
\usepackage{comment}
\usepackage[utf8]{inputenc} 
\usepackage[T1]{fontenc}    
\usepackage{hyperref}       
\usepackage{url}            
\usepackage{booktabs}       
\usepackage{mathtools}
\usepackage{amsfonts}   
\usepackage{amsmath}
\usepackage{nicefrac}       
\usepackage{microtype}      
\usepackage{graphicx}
\usepackage{subcaption}
\usepackage{algorithm}
\usepackage{xfrac}
\usepackage[noend]{algpseudocode}

\usepackage{adjustbox}
\usepackage{multirow}
\usepackage{makecell}

\usepackage{thmtools}
\usepackage{thm-restate}

\usepackage{xcolor}

\newcommand{\argmin}{\operatornamewithlimits{arg\,min}}
\newcommand{\argmax}{\operatornamewithlimits{arg\,max}}

\title{Brain-inspired Reverse Adversarial Examples}

\author{Shaokai Ye$^{1}$~~~Sia Huat Tan$^{1}$~~~Kaidi Xu$^{2}$~~~Yanzhi Wang$^{2}$~~~Chenglong Bao$^{1}$~~~Kaisheng Ma$^{1}$\\
 $^1$Tsinghua Univerity, China\\
    $^2$Northeastern University, USA\\ 
}

\begin{document}

\maketitle

\begin{abstract}

\input{abstract}

\end{abstract}

%\vspace*{-0.1in}
\input{intro}

\input{related.tex}

\begin{figure}[htb]
\centering
\vspace*{-0.15in}
\includegraphics[width=0.99\textwidth]{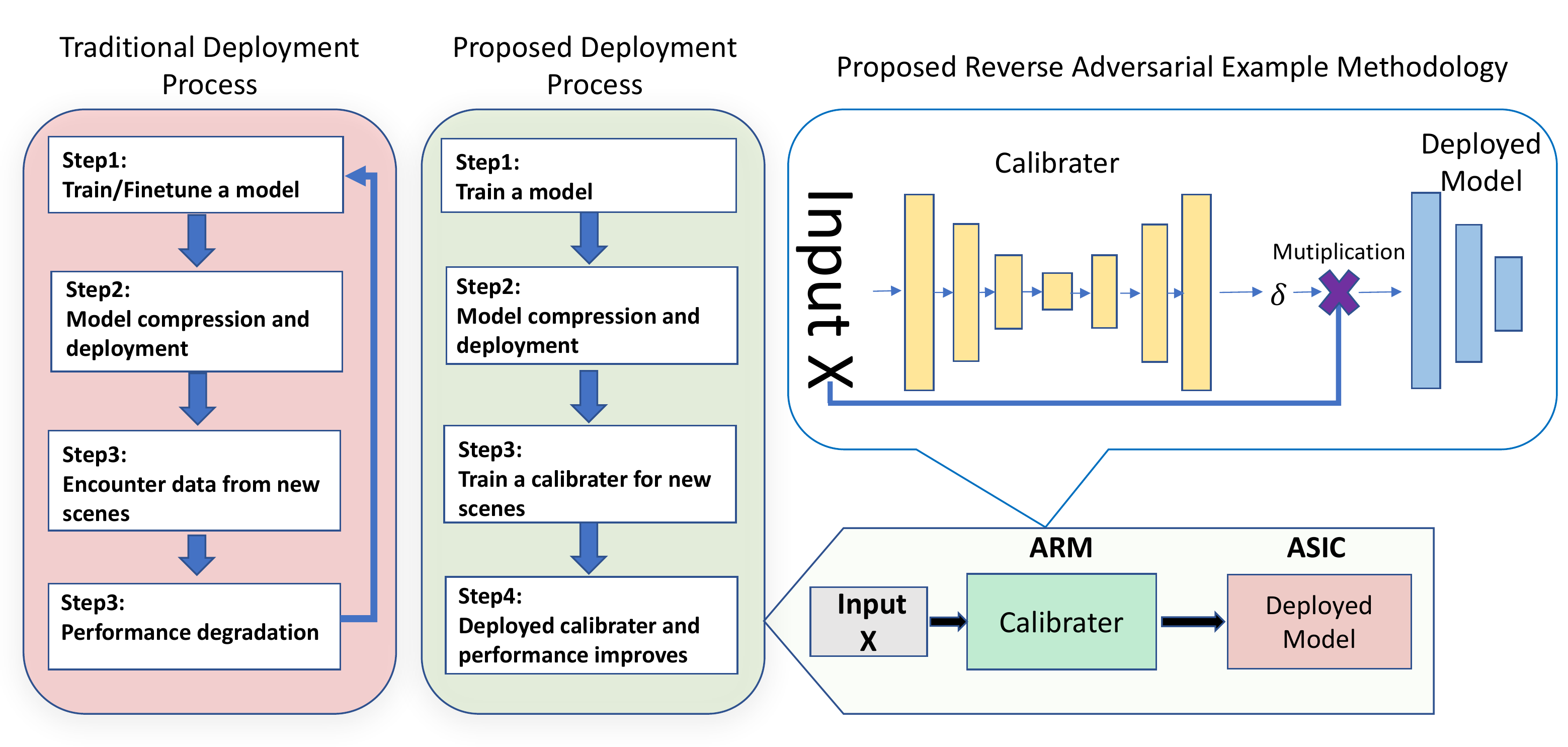}
\vspace*{-0.1in}
\caption{
Overview of proposed reverse adversarial example structure
}
\label{fig: Overview}
\end{figure}
\input{ProbStatement}

\input{Results.tex}

\input{Discussion}

\input{conclusion}

%\input{acknowledgements} 
% \small
% \bibliographystyle{plain}
% \bibliography{main}
\clearpage

%\appendix
%\normalsize
%\input{setup}
%\input{omitted_figures}
%\clearpage
%\input{new-proofs}
\end{document}

%% file: abstract.tex
A human does not have to see all elephants to recognize an animal as an elephant. On contrast, current state-of-the-art deep learning approaches heavily depend on the variety of training samples and the capacity of the network. In practice, the size of network is always limited and it is impossible to access all the data samples. 
Under this circumstance, deep learning models are extremely fragile to human-imperceivable adversarial examples, which impose threats to all safety critical systems. 
Inspired by the association and attention mechanisms of the human brain, we propose reverse adversarial examples method
that can greatly improve models' robustness on unseen data.
Experiments show that our reverse adversarial method can improve 
accuracy on average 19.02\% on ResNet18, MobileNet, and VGG16 on unseen data transformation. 
Besides, the proposed method is also applicable to compressed models and shows potential to compensate the robustness drop brought by model quantization - an absolute 30.78\% accuracy improvement.

%Therefore, inspired by the association mechanism of human brain, in this work, we show that by reversing the use of algorithm for generating adversarial examples, we are able to generate reverse adversarial examples that have positive effects. Those reverse adversarial examples can greatly improve performance of capacity limited models such as compressed models and signiﬁcantly improve models’ robustness against unseen data transformation. Experimental results demonstrate that our reverse adversarial methodology can improve as much as absolute 19.66% in accuracy on ResNet18, MobileNet and VGG16 in unseen data transformation. Besides, the proposed methodology is also applicable to compressed models and show potential to compensate the robustness drop brought by model quantization, a absolute 30.78% improvement in accuracy without unseen data transformation.

%% file: intro.tex
\section{Introduction~\label{section_introduction}}

During the process of the neural networks application systems being deployed, one problem that keeps standing out is systems' robustness and safety.
Restricted by robustness, many systems encounter difficulties during deployment or utilization.
For instance, autonomous vehicles - 
a fatal accident occured on a Tesla vehicle that is under autopilot mode, and the statement from Tesla says that "neither autopilot nor the driver noticed the white side of the tractor trailer against a brightly lit sky, so the brake was not applied"   \cite{autopilot}. In this case, the light condition can be seen as misleading perturbation to inputs that lead to the wrong decision in autopilot system. 
V2X (Vehicles to X) - for new camera devices added on the infrastructure, the model needs to be retrained from scratch to adjust to the light condition;
Smart shopping - a camera placed with light condition A needs to be retrained based on a model trained form data in light condition B.
Theoretically, if catastrophic forgetting can be overcome, 
robustness can be solved with enormous amount of data and
a network of an infinite volume. While in reality, most neural networks are deployed in scenarios with limited resources.
In order to better adapt to the limited computing resources, many techniques are even proposed to further reduce the volume - storage and computation - of the network, for instance, pruning~\cite{han2015learning,wen2016learning}, quantization~\cite{han2015deep,zhou2017incremental} and etc., which will further reduce the robustness of neural networks~\cite{guo2018sparse,lin2019defensive}.

Another critical issue for neural networks is the robustness for adversarial examples\cite{szegedy2013intriguing}. 
By adding maliciously crafted adversarial perturbation to inputs, attackers can make deep learning models mis-classify, even if the perturbation is imperceivable to human 
~\cite{carlini2017towards,goodfellow2014explaining,KurakinGB2016adversarial}.

In general, robustness and adversarial robustness are used to evaluate the capability of deep neural networks to accurately classify examples that are affected either by data transformation or maliciously crafted adversarial perturbation.

\begin{figure}[htb]
\centering
%\vspace*{-0.15in}
\includegraphics[scale=0.22]{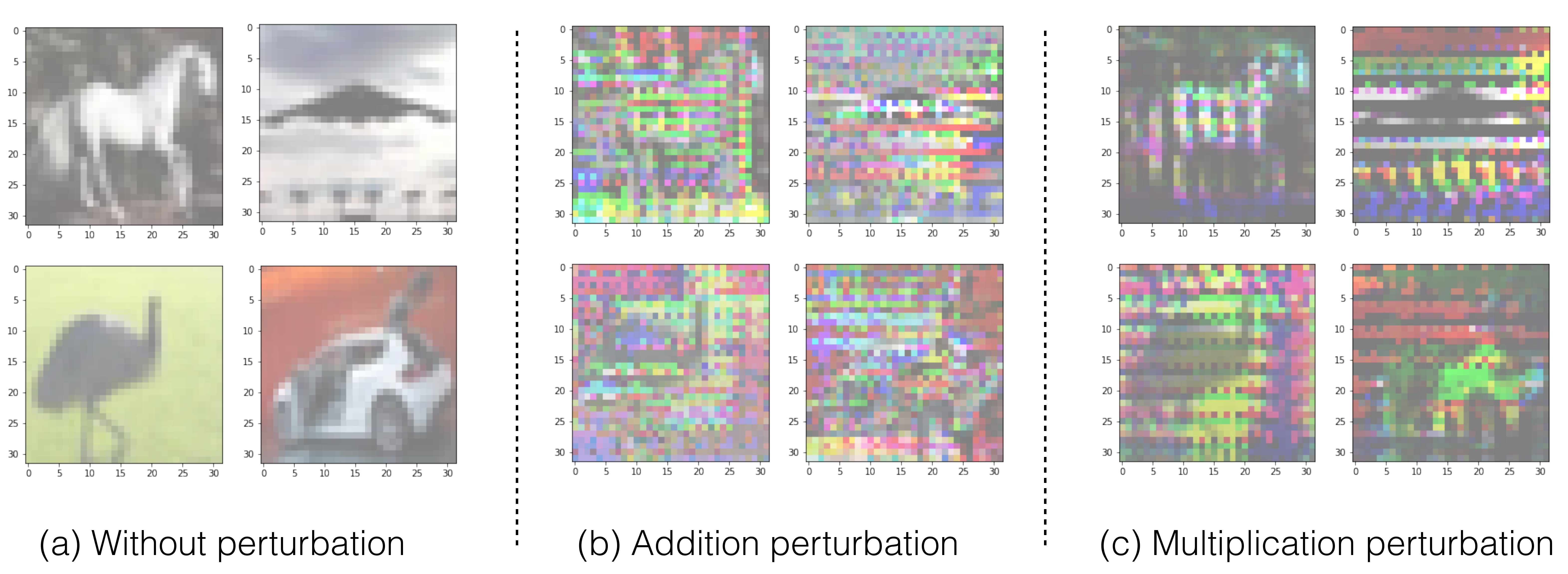}
%\vspace*{-0.1in}
\caption{
Visualization of two types of perturbations. (a) are original images. (b) are addition reverse adversarial perturbations. (3) are multiplication reverse adversarial perturbations. Compared to addition reverse adversarial perturbations, multiplication reverse adversarial perturbations carry more semantic meanings.
}
\label{fig: examples}
\end{figure}

\subsection{Motivations and Contributions~\label{brain-motivation1}}

\subsubsection{Brain-inspired robustness}

As to the aforementioned issues in neural networks, human brain is more robust.
From the perspective of psychology, there is a term called associationism~\cite{sep-associationist-thought} which refers to the association formed between the concepts of things. It can be classified as simple concept and complex concept. Two features of the associationism are contiguity which means that things or events with spatial or temporal proximity tend to be associated in the mind, and similarity which demonstrates that thought of one event tends to trigger the thought of a similar event. These characters are, to some extent, identical to what we have in system.
Inspired by associationism, we propose a calibrater
that can perform associationism-similar functions in neural networks.

In biology, all animals make decisions with extremely limited resources which are the energy extracted from food. In order to save energy for computation, the brain actually evolves a mechanism called attention~\cite{chen2012measuring,zhang2012neural,fang2012image}.
When a picture is reflected by eyes, only a small part of it is processed by the conscious processing system, while the rest are slightly computed and memorized by the sub-conscious system and some control-related-tasks are taken over by the muscle memory.
In this way, only the most important parts of the data are processed with extra attention.

Inspired by attention mechanism of human brain, we propose a 
multiplication based operation to merge benign perturbation generated by the calibrater to inputs.
Figure~\ref{fig: examples} shows the comparison between 
traditional addition method and attention-inspired multiplication method. 
%Figure~\ref{fig: examples}(a) are the original input images. Figure~\ref{fig: examples}(b) shows reverse adversarial perturbations that are merged into images through addition. 
%Figure~\ref{fig: examples}(c) are reverse adversarial perturbations that are merged with inputs with multiplication.
Note that these are pure perturbations, not perturbations merged into original inputs. Interestingly, the multiplication based perturbation carries more semantic meanings compared to addition based perturbation. This shows that the proposed way of merging perturbation creates perturbation that focuses more on modifying label-correlated area of the inputs.

\subsubsection{Existence of adversarial examples}

Deep learning models are so powerful in extracting
features from inputs, so that they are ultra sensitive to inputs.
Any alteration caused by either scenes or perturbations generated by adversarial attacks can result in severe impacts on the networks.
To overcome the problems caused by sensitivity of the networks,
the networks are trained against data with more complexity~\cite{goodfellow2014explaining,madry2017towards}.
We propose using a calibrater network that can capture
the connection between sets of data to improve the robustness.
If we view adversarial attacks as attacker PUSHING AWAY some input data from the true class to a target class (wrong for tasks human care), then it's likely that we can PULL input data TOWARDS the true class. In other words, we wish to optimize inputs 
such that they can be easily recognized by the given classifier.

As an extreme case study, a LeNet model is initialized with completely 
random weights, then tested on MNIST and CIFAR10 data set. The model is left untrained and we perform reverse adversarial attack by modifying previous adversarial attack algorithm \cite{KurakinGB2016adversarial} to craft reverse adversarial examples. Table~\ref{optimize_input} shows that we are able to achieve 87\% and 53\% accuracy without updating the randomized models. This example demonstrates that the inputs can be optimized towards given neural networks.

\begin{table}[htb]
\caption{
 \textbf{Results of randomly initialized weights under reverse adversarial attacks} (Acc. in $\%$)} 
  \label{optimize_input}
\centering
\begin{tabular}{c|c|c}
\toprule[1pt]
Model & \begin{tabular}[c]{@{}l@{}}Accuracy with an untrained \\randomly initialized network\end{tabular} & \begin{tabular}[c]{@{}l@{}}Accuracy of untrained models\\ under reverse adversarial attack\end{tabular} \\ \hline
LeNet on MNIST & 10.57 & 87.27 \\ \hline
LeNet on CIFAR10 & 10.31 & 53.38 \\ \bottomrule[1pt]
\end{tabular}
\end{table}

\textbf{Contributions:}
Our contributions are summarized as follows:

\begin{enumerate}
\item  To the best of our knowledge, this is the first attempt to discuss and utilize adversarial examples reversely. 
%The existence of reverse adversarial examples shows that it is possible to optimize inputs towards models.
Inspired by associationism, the proposed framework includes a generative network called calibrater to generate benign perturbation to pull the unlearnt features from 
unseen data to the learnt ones.
Furthermore, the multiplication based perturbation is imposed
to mimic the attention mechanism in the human brain,

\item  The results show that the proposed method can improve models' performance on unseen data with strong difference (data B2 in Table~\ref{data_transformation})
by 19.30\%, 18.11\%, 19.66\% for MobileNet, ResNet18, and VGG16 respectively. For compressed models (2 bits), 
models' performance can increase by 29.96\%, 28.76\%, and 20.67\%
on unseen data.

\item The proposed reverse adversarial examples method has broad potential applications.
Without retraining the main model, it can enhance scenario-specific robustness when the main model is deployed in ASIC for energy efficiency, and the calibrater is deployed in ARM Core for scenario-specific flexibility.

\end{enumerate}

%\item To the best of our knowledge, this is the first time that reverse adversarial examples are discussed and utilized. 
%Targeting at $W*X \rightarrow Y$, in traditional methods, $W$ is trained; 
%in generating adversarial examples, $X$ is added with fooling perturbation;
%while we show that for unseen $X'$, which is not fully independent and identically distributed with $X$,
%it is possible to add benign perturbation $P$ on $X'$
%to make $W*(P*X')$ more close to $Y$.
%Compared to adversarial examples, reverse adversarial examples can create benign perturbation to improve models robustness.

%% file: related.tex
\section{Related work}

\textbf{Adversarial attacks.} 
Adversarial examples~\cite{szegedy2013intriguing,biggio2013evasion} are created by adding human-imperceivable perturbation to inputs. These malicious examples are transferable \cite{papernot2016transferability,papernot2017practical,liu2016delving,xu2018structured}, and are even effective in physical world~\cite{athalye2017synthesizing,song2018physical}.

There are many methods for generating adversarial examples, such as FGSM~\cite{goodfellow2014explaining}, IFGSM~\cite{KurakinGB2016adversarial} and C\&W~\cite{carlini2017towards}.  
However, adversarial examples have values beyond malicious attacks. Inspired by the vulnerable nature of neural networks, the calibrater we proposed is able to perform reverse adversarial attacks so that powerful attack methods can turn to benefit neural networks.

\textbf{Adversarial training.} Goodfellow \emph{et al.}~\cite{goodfellow2014explaining} train a mixture of nature examples and adversarial examples to improve models' robustness. Madry \emph {et al.}~\cite{madry2017towards} formulate adversarial training as a min-max optimization problem and solve it iteratively, which is the only method that passes the obfuscated gradient check~\cite{athalye2018obfuscated}.

Those methods either partly or fully use adversarial examples as training data, resulting in models that are robust to adversarial examples, but along with cost of accuracy degradation in nature examples. In resources constrained systems, requiring a small model to train adversarial examples (or a mixture of nature and adversarial examples) is not feasible.

\textbf{Generative networks for adversarial examples.} 
Prior works~\cite{xiao2018generating,hu2017generating} use generative models to produce adversarial examples. Those work are able to generate effective adversarial examples in feed-forward process.
However, those methods only focus on generating malicious perturbation for attacks, rather than generating reverse adversarial examples for benign uses.

\begin{comment}
\textbf{Model Compression.} 
In order to let large models to be deployed on resources-constrained edge systems, many model compression methods are usually adopted to reduce both the storage and computation cost,
for example, weight pruning\cite{han2015learning}, weight quantization ~\cite{zhou2017incremental,han2015deep} and model distillation are proposed to compress neural networks %~\cite{buciluǎ2006model,furlanello2018born,hinton2015distilling}.
However, these methods often require retraining, which is not feasible in many application scenarios.
\end{comment}

%% file: ProbStatement.tex
\section{Proposed brain-inspired calibrater}

One common deployment process for deep learning models usually starts with training a large model in one specific training dataset, which is assumed to be close to the applications. Once the model is trained, it is deployed in hardwares followed by the model compression on the same training set. However, the real world is complicated - the assumption that the training data and data in deployment are independently and identically distributed can seldomly hold: scenes might change. For example, in self-driving training, the data collected in a city that is always sunny is very different from those of a city that rains a lot.
In such cases, the performance degradation occurs. 
In order to compensate the performance degradation, the traditional approach is either adding the new data to fine-tune the original model or training a completely new model that is only good for the new scene (see the left part of figure \ref{fig: Overview}). In both cases, the training cost of the main model is huge and this issue scales up quickly when the size of the main model and the number of deployed devices are increased.

%Figure \ref{fig: Overview} shows an overview of the proposed framework. Traditional deployment process for deep learning models usually starts with training a model in one specific training dataset, which is assumed to be close to the application scenes. Once the model is trained, a common practice is to use the same dataset for compression training. After that, the model is deployed in hardwares such as ASIC. However, the real world is complicated - the assumption that the the training data and data in deployment are independent and identically distributed can seldom hold: scenes might change: for self-driving data collected in a city that is always sunny is very different from those of a city that rains a lot.
%In such cases, the performance degradation occurs. 
%In order to compensate the performance degradation, the traditional approach is either adding the new data to fine-tune the original model or training a completely new model that is only good for the new scene, in both cases the training cost of the main model is huge. This issue scales up quickly with the size of main model and the deployment amount of devices.

The proposed framework approaches to this problem in a very different way: instead of asking models to fit every possible scene, it's more feasible to train a calibrater with new data and help the main model understand data from new scenes.
The middle part of Figure~\ref{fig: Overview} demonstrates our proposed deployment process. 
Instead of the training the main model again, we propose to train a light weight generative calibrater when the additional unseen data have been applied. Compared to the traditional deployment process, our approach significantly accelerates the deployment process. Meanwhile, the proposed generative calibrater is feasible to run in an ARM embedded processor due to its simplicity. Once the deployed device is switched for different scenarios, it is very flexible to update the calibrater running on ARM, rather than the main model running on ASIC. In the following context, we introduce the details in training our proposed calibrater.

\subsection{Brain-inspired calibrater}
In this section, we define $f(\cdot,\theta)$ to be a neural network with parameters $\theta$ and $c(x) = \argmax(f(x))$ to be the output classification  result of the neural network. 
Given a training set $\mathcal{X}_1 = \{(x_i,y_i)\}_{i=1}^n$, the network parameter $\theta$ is obtained via solving the minimization:
\begin{equation}\label{eq:train}
\theta^* \in \argmin_{\theta}\frac{1}{n}\sum_{i=1}^n\mathrm{loss}(f(x_i,\theta),y_i),
\end{equation}
where $\mathrm{loss}$ is the loss function defined by users. For a given network $f(x,\theta^*)$, typical adversarial attack methods find the perturbation $\delta$ via the optimization model~\cite{goodfellow2014explaining,KurakinGB2016adversarial,carlini2017towards}:
\begin{equation}\label{adsam}
\begin{aligned}
\min_{\delta} \|\delta\|,~\mbox{s.t.}~  c(f(x+\delta,\theta^*))\neq c(f(x,\theta^*))\mbox{ or }
\max_{\delta} \ell(f(x+\delta,\theta^*),f(x,\theta^*)), ~ \mbox{s.t.}~ \|\delta\|\leq\Delta,
\end{aligned}
\end{equation}
where $\ell$ measures the difference between the network prediction of $x+\delta$ and $x$ and $\Delta$ is the perturbation range. It is noted that the adversarial attacks always exists for a given classifier, i.e.\ there exists bad samples in the local region of the given data. Motivated by this observation, we propose to reverse the process of generating adversarial attacks such that to find a better perturbation of the given sample which can be easily recognized by the trained classifier. More concretely, given a set of unseen data $\mathcal{X}_2=\{(x_i^{'},y_i^{'})\}_{i=1}^m$ where $\mathcal{X}_1\cap\mathcal{X}_2=\emptyset$, denote the benign perturbation by $G(x,\theta_G)$ where $\theta_G$ are the parameters of $G$, we aim at finding the perturbation $\delta=G(x,\theta_G)$ via minimizing the loss
\begin{equation}\label{eq:fx}
\small
%L(\mathbf{x}_i\cdot G(\mathbf{x}_i),\mathbf{y}_i,\theta_{G}) = 1 - F(\mathbf{x}_i \cdot  G(\mathbf{x}_i),\theta_{M})_{t_i} ,
\min_{\theta_G}~L_{G} = \frac{1}{m}\sum_{i=1}^m\mathrm{loss}(f(x_i^{'}\odot G(x_i^{'},\theta_{G}), \theta^*),y_i^{'}),
\end{equation}
where $\odot$ is the point-wise multiplication. Compared to the adversarial attack \eqref{adsam}, our calibrator defers from two perspectives: (i) we perform reverse adversarial attack by requiring the generative model to output $\delta$ to minimize classification loss of the main model $f({x}\odot\delta)$ in the new dataset $\mathcal{X}_2$; (ii) instead of the addition of the perturbation in \eqref{adsam}, point-wise multiplication mimics the attention mechanism in brain as shown in prior work~\cite{chen2012measuring,zhang2012neural,fang2012image}. 

\begin{algorithm}[h]  
\caption{Training for proposed calibrater $G$}  
\begin{algorithmic}[1]  
%\Require{Dataset A, main model M, dataset B, generative model G, optimizer Opt}
%\Input{Dataset $\mathcal{X}_1$ and $\mathcal{X}_2$}
%\Output{Generative model $G$}
\State{{\bf Inputs:} Dataset $\mathcal{X}_1$ and $\mathcal{X}_2$, the main classifier $f(\cdot,\theta)$ and calibrater $G(\cdot,\theta_G)$}
\State{{\bf Outputs:} $\theta^*$ and $\theta_G^*$}
\State{Train the main classifier $f(\cdot,\theta)$ in dataset $\mathcal{X}_1$ via solving \eqref{eq:train}}.

 \State {// \emph{Train the calibrater $G(\cdot,\theta_G)$ in dataset $\mathcal{X}_2$ via solving \eqref{eq:fx}}}
\For{batch in Dataloader(dataset $\mathcal{X}_2$)}
\State{$data,target = getData(batch)$ // \emph{          get data and true labels in a batch}}
\State{$\delta = G.getOutput(data)$ // \emph{get output from the calibrater}}
\State{$F = f(\cdot,\theta).getSoftMaxOutputs(data \cdot \delta)$ // \emph{get softmax output from the classifier}} 
\State{$index = argmax(target)$ // \emph{get the index of where the true label has maximal value 1}}
\State{$L_{G} = 1- F[index]$ // \emph{add loss if the softmax output of classifier on that index is away from 1 }} 
\State{$grad = Opt.getGradient(L_{G})$ // \emph{calculate the gradients for G based on the loss}}
\State{$weightUpdate(G,grad)$ // \emph{update network G based on the gradients}}
\EndFor
\end{algorithmic}  
\label{train}
\end{algorithm}

%% file: Results.tex
\section{Results}

\subsection{Experiment setup}
In this section, we evaluate the proposed benign perturbation calibrater based on CIFAR10 data set, using MobileNet~\cite{howard2017mobilenets}, VGG16~\cite{simonyan2014very}, and ResNet18~\cite{he2016deep} models (also referred as main models). Our implementation is based on PyTorch and we use 4 different data augmentation functions in PyTorch to define the so called A/B scenario setting. We first assume that training data and test data for CIFAR10 are identically and independently distributed. We refer data that are transformed via random crop resize and random horizontal flip as scenario A, and refer data that are transformed via random rotation and random color jittering as scenario B. For further comparison, we also derive B1 and B2 as scenario B under increasing strength of data transformations. For scenario B1, maximum rotation is 15, maximum change of brightness is 0.8, maximum change of contrast is 0.8, maximum change of saturation is 0.8. And for scenario B2, the maximum rotation is 20, maximum change of brightness is 2, maximum change of contrast is 2 and maximum change of saturation is 2. For scenario A, we only use PyTorch's default settings.
All main models are trained under scenario A in training data, and tested under scenario B1/B2 in test data. The calibraters are trained under scenario B1/B2 in training data and used to support main models under scenario B1/B2 in test data.
The backbone of the generative model is consisted
of 3 down-sampling convolutional layers, followed by a number of residual blocks and 3 up-sampling convolutional layers.
The sizes of the calibraters are chosen to be one tenth of the main models. To control the size of the calibrater, we vary the number of channels within residual blocks and number of residual blocks. For example, to have a calibrater that has one tenth parameter the size of MobileNet, we use 2 residual blocks with 18 channels.
The solver chosen for training main models and calibraters is ADAM\cite{KingmaB2015adam}. Training for calibraters takes 200 epochs and the initial learning rate is set to be 0.0002 before 50 epochs, 0.0001 at 50-100 epochs, 0.00005 at 100-150 epochs, and 0.00002 at 150-200 epochs. During training of calibraters, main models are only used for feed-forward. 
\subsection{Robustness improvement brought by reverse adversarial examples method}

\textbf{Results on uncompressed models:}~Table~\ref{data_transformation} shows the robustness increase
brought by the proposed method. In Table~\ref{data_transformation}
Column 3 lists models' test accuracy on clean test data without any data augmentation. When models are tested under scenario B1 and B2 in test data, models' accuracy drop about 20\% to 40\%, as shown in  
Column 4 in Table~\ref{data_transformation}). We consider random rotation and random color jittering as common environmental factors in real world and this result demonstrates that deep learning models lack of robustness in our settings.
Column 5 shows that by using the proposed calibrater, we can achieve 4.8\% to 19.66\% accuracy boost (Column 5 in Table~\ref{data_transformation}), with only little overhead brought by the calibrater (Column 6\&7 in Table~\ref{data_transformation}).

\begin{table}[]
\small
\caption{
 \textbf{Robustness improved by proposed method} (Accuracy in $\%$) on Cifar10} 
  \label{data_transformation}
%\resizebox{\textwidth}{!}{%
\centering
\begin{tabular}{l|l|l|l|l|l|l}
\toprule[1pt]
\begin{tabular}[c]{@{}l@{}}Test\\Domain\end{tabular} & Networks & \begin{tabular}[c]{@{}l@{}}Ideal Case\\Accuracy:\\Training\\on A+\\Test on A\end{tabular} & 
\begin{tabular}[c]{@{}l@{}}Baseline\\Accuracy:\\Training on A\\No Calibrater\\Test on B1/B2 \end{tabular} & 
\begin{tabular}[c]{@{}l@{}}Accuracy:\\Training on A \\Calibrater-mul.\\ on B1/B2
\\Test on B1/B2 \end{tabular} & \begin{tabular}[c]{@{}l@{}}Model\\Size\\(MB)\end{tabular} & \begin{tabular}[c]{@{}l@{}}Calibrater\\Size \\(MB)\end{tabular} \\ \hline
\multirow{3}{*}{\begin{tabular}[c]{@{}l@{}}Test on\\ B1*\end{tabular}} & MobileNet & 90.88 & 65.78  & \textbf{75.59(+9.81)} & 12.88 & 1.24 \\ %\cline{2-7} 
 & ResNet18 & 93.96 & 73.84  & \textbf{78.62(+4.78)} & 44.80 & 4.32 \\ %\cline{2-7} 
 & VGG16 & 91.61 & 70.11  & \textbf{80.82(+10.71)} & 58.80 & 5.48 \\ \hline
\multirow{3}{*}{\begin{tabular}[c]{@{}l@{}}Test on\\ B2$^\#$\end{tabular}} & MobileNet & 90.88 & 43.78  & \textbf{63.08(+19.30)} & 12.88 & 1.24 \\ %\cline{2-7} 
 & ResNet18 & 93.96 & 48.35  & \textbf{66.46(+18.11)} & 44.80 & 4.32 \\ %\cline{2-7} 
 & VGG16 & 91.61 & 48.38 & \textbf{68.04(+19.66)} & 58.80 & 5.48 \\ 
 \bottomrule[1pt]
\end{tabular}%
%}

\flushleft

+ A: Random crop resize and random horizontal flip.

* B1: Maximum rotation is 15 degree, maximum change of brightness, contrast, saturation are 0.8.

$^\#$B2: Maximum rotation is 20 degree, maximum change of brightness, contrast, saturation are 2.
\end{table}

\begin{table}[]
\small
\centering
\caption{
 \textbf{Robustness improvement on compressed models} (Accuracy in $\%$) on Cifar10} 
  \label{model_compression}
\begin{tabular}{l|l|l|l|l|l|l}
\toprule[1pt]
\begin{tabular}[c]{@{}l@{}}Number \\of bits\end{tabular} & Networks & \begin{tabular}[c]{@{}l@{}}Quantized models \\Acc. on testset\\ without \\data augmentation\end{tabular} & \begin{tabular}[c]{@{}l@{}}Baselines: \\Quantized\\Models \\ Acc. on B2 \end{tabular} & \begin{tabular}[c]{@{}l@{}} Quantized \\Model\\ Acc. on B2 \\with Calibrater\end{tabular} & \begin{tabular}[c]{@{}l@{}}Model\\Size\\ (MB)\end{tabular} & \begin{tabular}[c]{@{}l@{}}Calibrater\\Size\\ (MB)\end{tabular} \\ \hline
\multirow{3}{*}{3 bits} & MobileNet & 83.69 & 35.63 & \textbf{66.33(+30.70)} & 3.22 & 0.31 \\ %\cline{2-7} 
 & ResNet18 & 90.00 & 39.41 & \textbf{70.18(+30.77)} & 11.20 & 1.08 \\ %\cline{2-7} 
 & VGG16 & 91.53 & 46.69 & \textbf{67.46(+20.77)} & 14.70 & 1.37 \\ \hline
\multirow{3}{*}{2 bits} & MobileNet & 77.96 & 34.33 & \textbf{64.29(+29.96)} & 1.21 & 0.31 \\ %\cline{2-7} 
 & ResNet18 & 89.87 & 41.4 & \textbf{70.16(+28.76)} & 4.20 & 1.08 \\ %\cline{2-7} 
 & VGG16 & 89.11 & 45.44 & \textbf{66.11(+20.67)} & 5.51 & 1.37 \\ \bottomrule[1pt]
\end{tabular}%
%}
\end{table}

\subsection{Robustness improvement on compressed models}
\textbf{Results on compressed models:}~In this subsection, we demonstrate the effectiveness of the proposed calibrater on compressed models. Main models' weights are quantized to 2/3 bits using deep compression\cite{han2015deep}. To ensure that overhead brought by our calibraters is still low compared to the main models, we quantize them to 8 bits without any quantization training. Note that the calibraters' training can naturally leverage the accelerated inference speed of main models. In comparison between Column 4 in Table \ref{data_transformation} and Column 3 in Table~\ref{model_compression}, we observe that quantized models suffer significantly more accuracy drop under scenario B2 in test data. We consider this phenomenon as an indicator that model compression hurts models' robustness in our settings.  Table~\ref{model_compression} Column 5 shows that when deployed the calibraters, the models' performance on B2 improves significantly. 

%\emph{Compression does harm robustness.}
%Alough VGG has been proposed for several years, among all the 
%three networks we evaluated, it has the best robustness after quantization,
%which further proves the work proposed by \Note{xx et al.}
%Traditional quantization targets for on accuracy comparison on same training and test A, making accuracy in
%Column 3 in Table~\ref{model_compression} more close to olumn 3 in Table~\ref{data_transformation}).
%While we suggest evaluation metrics with both accuracy on
%same data, and robustness.

%% file: Discussion.tex
\section{Discussion}
\begin{figure}[htb]
\centering
\vspace*{-0.15in}
\includegraphics[width=1\textwidth]{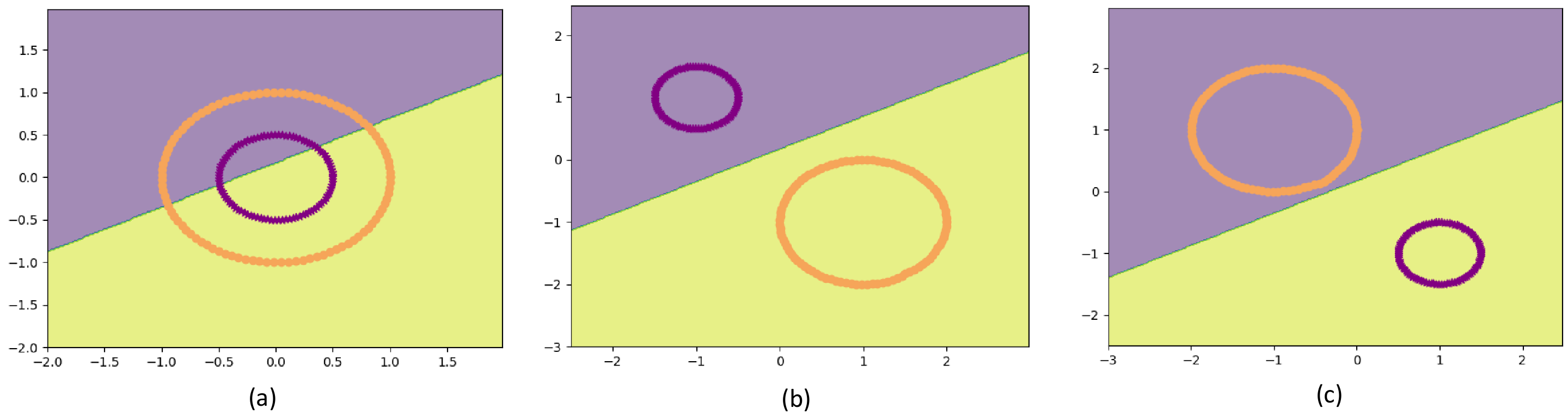}
\vspace*{-0.15in}
\caption{
Visualization that shows adversarial attacks and reverse adversarial attacks are symmetric. 
(a), Results of a linear classifier - a linear classifier's decision boundary on non-linear-separable data.
(b), Results by reverse adversarial examples - the benign perturbation moves inputs to correct regions of the linear classifier's decision boundary.
(c), Results by adversarial examples - data are moved to wrong regions, resulting in 100\% misclassification, similar to adversarial attack.
}
\label{linear}
\end{figure}

\begin{table}[]
\caption{
 \textbf{Transferability test} (Accuracy in $\%$) on Cifar10 on B2} 
  \label{transferability}
\centering
\begin{tabular}{c|c|c|c}
\toprule[1pt]
\begin{tabular}[c]{@{}l@{}}Tested on\\ \textbackslash{}Calibrater trained on B2\end{tabular} & MobileNet & ResNet18 & VGG16 \\ \hline
MobileNet(43.78) & \textbackslash{} &56.5(+12.72)  & 51.61(+7.83) \\ \hline
ResNet18(48.35) &64.13(+15.78)  & \textbackslash{} & 61.42(+13.07) \\ \hline
VGG16(48.38) & 58.27(+9.89) & 57.58(+9.19) & \textbackslash{} \\
\bottomrule[1pt]
\end{tabular}
\end{table}

\subsection{The symmetry between adversarial attacks and reverse adversarial attacks}
Using 2-dimensional synthetic data as an example, we visualize the
proposed method and demonstrate that adversarial examples and reverse adversarial examples are essentially symmetric. Note that in this subsection, we use addition based perturbation to help demonstrate the symmetry phenomenon.
In Figure~\ref{linear}(a), it can be observed that the synthetic data form two concentric circles that represent two classes of data that are not linear separable. A linear classifier is applied to do classification on data, and the plotted decision boundary shows that the linear classifier is not able to fully separate data.
In Figure~\ref{linear}(b), we train a calibrater on these data and generate benign perturbation to help the linear classifier do classification. It is clearly observed that after perturbation are applied on input data, data are moved to different regions and become linear separable. In this example, we show that the calibrater is able to optimize inputs towards the linear classifier's decision boundary,  solving a problem that is impossible for a linear classifier in an elegant way.
In Figure~\ref{linear}(c), the calibrater has been used reversely.
It is noted that data are moved to opposite locations compared to Figure~\ref{linear}(b), demonstrating that the reverse use of our calibrater are able to perform adversarial attacks. This example shows that adversarial attacks and reverse adversarial attacks might be symmetric, as attacks harm the performance of models and reverse attacks benefit models.

\subsection{Proposed calibrater improves models' robustness against unseen data}
\begin{figure}[ht]
\vspace*{-0.1in}
\centering
\includegraphics[scale = 0.73]{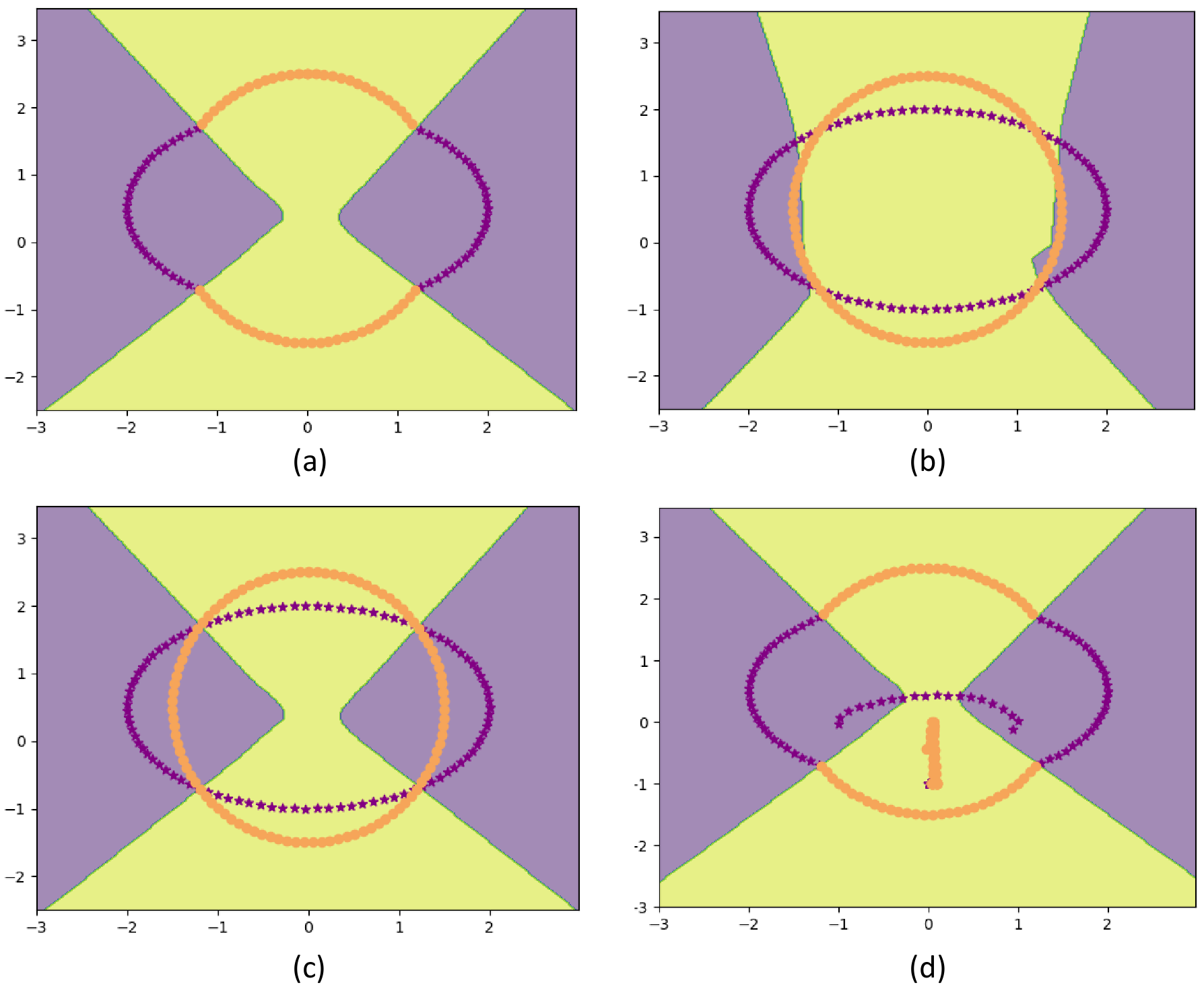}
\vspace*{-0.1in}
\caption{
Visualization of how the calibrater handles unseen data.
(a), MLP achieves 100\% classification on partial data, depicted by dots that are inside decision boundary(Note the difference between (a) and (c)). 
(b), Results by a MLP trained with full data. The accuracy is is 63\% for the whole data.
(c), Based on the main model in (a), data hidden in (a) are used to train a calibrater.
(d), Results by the main model trained in (a) and the calibrater.
Test accuracy on excluded data by the main model is 68.7\%.
For comparison, for unseen data, (a) accuracy is 0\%, and (b)
accuracy is 18.75\%.
Benign perturbation move the previously excluded data towards to MLP's decision boundary.
}
\label{nonlinear}
\end{figure}

In Figure~\ref{nonlinear}(a), we create synthetic data that form two overlapping ellipses (data are those orange dots and purple dots instead of the regions). Figure~\ref{nonlinear}(c) helps to understand our setting: data from both ellipses form what can be seen as overlapping region. In this subsection, data that are on the non-overlapping region are chosen as training data for the main model, and data on overlapping regions are chosen as training data for the calibrater.
Here, a MLP(Multiple Layer Perceptron) is used as the main model to classify its training data. Not surprisingly, it achieves 100\% accuracy as shown in Figure~\ref{nonlinear}(a). However, if we ask MLP to classify data that were hidden from its training, the resulted accuracy is as low as 0\%.
This example represents what commonly happens in real world scenarios.
In practice, as shown in Figure~\ref{nonlinear}(b), once the unseen data are realized, the model is put to finetune against those added data. Even in this costly way, the model can only achieve 18.75\% accuracy in the added data. On contrast, Figure~\ref{nonlinear}(d) shows that if the calibrater is applied, without needing to finetune the main model, the calibrater is able to 
transfer the unseen data so as to fit the trained main model.
As a result, the main model achieves 68.7\% accuracy in data that was unseen by the main model.
This example shows the true power of the proposed method: the deployment of calibrater greatly improves models' robustness against unseen data and reduces the deployment cost.
A deep learning model deployed with the calibrater does not have to repeat the costly recycle process in the present of data from new scenes, which makes it possible to deploy neural networks in a completely new fashion.

\subsection{Reverse adversarial examples are transferable}

So far, all experiments are performed in a way that every calibrater is trained with a main model it targets for . It's natural to raise a question whether it's necessary to train a calibrater for every main model individually.
In Table \ref{transferability}, we show that just like adversarial examples that are transferable\cite{papernot2016transferability,papernot2017practical,liu2016delving}, reverse adversarial examples are transferable as well. Columns in Table \ref{transferability} list main models used for training the calibraters and the rows list the main models used for testing their adaptivity with calibraters trained with models from quite different architectures. Here, we keep the A/B setting and test models' performance under scenario B2 in the test set of CIFAR10. The results show that all models' performance improve significantly with the use of the calibrater, regardless of which models the calibraters are trained with.

%% file: conclusion.tex
\section{Conclusion}

In this work, we show that it is possible to create reverse adversarial examples and learn a generative model called calibrater to construct them. Unlike adversarial examples that are mainly used for attacking deep learning models, reverse adversarial examples can significantly improve models' robustness in unseen data transformations, which assembles human brain that associates new knowledge with old knowledge. Furthermore, we borrow the idea of attention mechanism in human brain and introduce attention-inspired multiplication perturbation, which carries more semantic meaning. The proposed method takes a novel perspective to tackle problems frequently encountered in the deployment of neural networks.